\documentclass[runningheads, envcountsame, a4paper]{llncs}

\usepackage{amsmath,amsfonts,amssymb}
\usepackage[ruled,linesnumbered,boxed]{algorithm2e}
\usepackage{mwe,graphicx,tabu,subcaption,float}
\usepackage{color,soul}
\usepackage{float}
\usepackage{tcolorbox}
\usepackage{hyperref}
\SetKwComment{Comment}{$\triangleright$\ }{}

\SetKwInOut{Parameter}{Parameters}
\ifdefined\paperDraft
 \def\mycomment#1{{\color{blue}[\textit{#1}]}}
 
 \else
 \def\mycomment#1{}
 
\fi

\begin{document}
\title{Language comparison via network topology}
%
%
\author{Bla\v{z} \v{S}krlj\inst{1,2} \and
Senja Pollak\inst{2,3}}
\authorrunning{\v{S}krlj, B. and Pollak, S.}
%
\institute{Jo\v{z}ef Stefan International Postgraduate School \and
Jo\v{z}ef Stefan Institute, Slovenia \and 
Usher Institute, Medical School, University of Edinburgh, Edinburgh\\
\email{\{blaz.skrlj,senja.pollak@ijs.si\}}}
\maketitle              
\begin{tcolorbox}
 The final authenticated publication is available online at \url{10.1007/978-3-030-31372-2_10}.
\end{tcolorbox}

\begin{abstract}
Modeling relations between languages can offer understanding of language characteristics and uncover similarities and differences between languages. Automated methods applied to large textual corpora can be seen as opportunities for novel statistical studies of language development over time, as well as for improving cross-lingual natural language processing techniques. In this work, we first propose how to represent textual data as a directed, weighted network by the text2net algorithm. We next explore how various fast, network-topological metrics, such as network community structure, can be used for cross-lingual comparisons. In our experiments, we employ eight different network topology metrics, and empirically showcase on a parallel corpus, how the methods can be used for modeling the relations between nine selected languages. We demonstrate that the proposed method scales to large corpora consisting of hundreds of thousands of aligned sentences on an of-the-shelf laptop. We observe that on the one hand properties such as communities, capture some of the known differences between the languages, while others can be seen as novel opportunities for linguistic studies.

\keywords{Computational typology \and cross-linguistic variation \and network theory \and language modeling \and comparative linguistics \and graphs \and language representation}

\end{abstract}

\section{Introduction and related work}
Understanding cross-linguistic variation has for long been one of the foci of linguistics, addressed by researchers in comparative linguistics, linguistic typology and others, who are motivated by comparison of languages for genetic or typological classification, as well as many other theoretical or applied tasks. Comparative linguistics seeks to identify and elucidate genetic relationships between languages and hence to identify language families \cite{TraskDicto}. From a different angle, linguistic typology compares languages to learn how different languages are, to see how far these differences may go, and to find out what generalizations can be made regarding cross-linguistic variation on different levels of language structure and aims at mapping the languages into types \cite{Daniel2010}. The availability of large electronic text collections, and especially large parallel corpora, have offered new possibilities for computational methodologies that are developed to capture cross-linguistic variation. This work falls under computational typology \cite{kepser2008linguistic,asgari2017past}, an emerging field with the goal of understanding of the differences between languages via computational (quantitative) measures. Recent studies already offer novel insights into the inner structure of languages with respect to various sequence fingerprint comparison metrics, such as for example the Jaccard measure, the intra edit distance and many other boolean distances \cite{rama2012good}. Such comparisons represent 
e.g., sentences as vectors, and evaluate their similarity using plethora of possible metrics. Albeit useful, vector-based representation of words, sentences or broader context does not necessarily capture the context relevant to the task at hand and the overall structure of a text collection. Word or sentence embeddings, which recently serve as the language representation workhorse, are not trivial to compare across languages, and can be expensive to train for new languages and language pairs (e.g., BERT \cite{devlin2018bert}).
Further, such embeddings can be very general, possibly problematic for use on smaller data sets and are dependent on input sequence length. 

In recent years, several novel approaches to computational typography have been applied. For example, Bjerva et al. \cite{bjerva2019language} compared different languages based on distance metrics computed on universal dependency trees \cite{nivre2016universal}. 
They discuss whether such language representations can model geographical, structural or family distances between languages. Their work shows how a two layer LSTM neural network \cite{hochreiter1997long} represents the language in a structural manner, as the embeddings mostly correlate with structural properties of a language. Their main focus 
is thus on explaining the structural properties of neural network word embeddings. Algebraic topology was also successfully used to study syntax properties by Port et al. \cite{port2018persistent}.  
Similar efforts of statistical modelling of language distances were previously presented in e.g., \cite{kettunen2006analysis} who used Kolmogorov complexity metrics.

In contrast, we propose a different approach to modeling language data. The work is inspired by ideas of node representation as seen in contemporary geometric and manifold learning \cite{goyal2018graph} and the premises of computational network theory, which studies the properties of interconnected systems,  found within virtually every field of science \cite{zhang2018network}. Various granularities of a given network can be explored using approaches for community detection, node ranking, anomaly identification and similar \cite{fortunato2010community,kralj2019netsdm,brandao2017social}. We demonstrate that especially information flow-based community detection \cite{de2015identifying} offers interesting results, as it directly simulates information transfer across a given corpus. In the proposed approach, we thus model a corpus (language) as a single network, exposing the obtained representation to powerful network-based approaches, which can be used for language comparison (as demonstrated in this work), but also for e.g., keyword extraction (cf. \cite{boudin:2016:COLINGDEMO} who used TopicRank) 
and potentially also for representation learning and end-to-end classification tasks.

The purpose of this work is twofold. First, we explore how a text can be transformed into a network with minimal loss of information. We believe that this powerful and computationally efficient text representation that we name
 \emph{text2net}, standing for text-to-network transformation,
can be used for many new tasks. Next, we show how the obtained networks can be used for cross-lingual analysis across nine languages (36 language pairs).
 
This work is structured as follows. In Section~2 we introduce the networks and the proposed text2net algorithm.
Next, we discuss network-topological metrics (Section~\ref{sec:metrics}) that we use for the language comparison experiment in Section~\ref{sec:comparison}. The results are presented in Section~\ref{sec:five}, followed by discussion and conclusions in Section~\ref{sec:disc}.
\section{Network-based text representation}
First, we discuss the notion of networks, and next present our 
text2net approach.

\subsection{Networks}
\label{sec:net}
We first formally define the type of networks considered in this work.

\begin{definition}[Network]
A network is an object consisting of nodes, connected by arcs (directed) and /or edges (undirected). In this work we focus on directed networks, where we denote with $G = (N,A)$ a network $G$, consisting of a set of nodes $N$ and a set of arcs $A \subseteq N \times N$ (ordered pairs).
\end{definition}

Such simple networks are not necessarily informative enough for complex, real world data. Hence, we exploit the notion of weighted directed networks.

\begin{definition}[Directed weighted network]
A directed weighted network is defined as a directed network with additional, real-valued weights assigned to arcs.
\end{definition}


Note that assigning weights to arcs has two immediate consequences: arcs can easily be pruned (using a threshold), and further, algorithms, which exploit arc weights can be used.
We continue to discuss how a given text is first transformed into a directed weighted network $G$. 

\subsection{text2net algorithm}
\label{sec:text2net}
Given a corpus $T$, we discuss the mapping $\textrm{text2net}: T \rightarrow G$. As text is sequential, the approach captures global word neighborhood, proceeding as follows:

\begin{enumerate}
\item Text is first tokenized and optionally stemming, lemmatization and other preprocessing techniques are applied to reduce the space of words.
\item text2net traverses each input sequence of tokens (e.g., words, or lemmas or stems depending on Step 1), and for each token (node) stores its successor as a new node connected with the outbound arc. This step can be understood as breaking the the text into triplets, where two consecutive words are connected via a directed arc (therefore preserving the sequential information). 
\item During construction of such triplets, arcs commonly repeat, as words often appear in same order. Such repetitions are represented as arc weights. Weight assignment can depend on the arc type. For this purpose, we introduce a mapping $\rho(a) \rightarrow \mathbb{R}; a \in A$ ($A$ is the set of arcs), a mapping which assigns a real value to a given arc with respect to that arc's properties.
\item Result is a weighted, directed network representing weighted token co-occurrence.
\end{enumerate}

The algorithm can thus formally be stated as given in Algorithm~\ref{algo:plab}. 
The key idea is to incrementally construct a network based on text, while traversing the corpus \emph{only once} (after potential selected preprocessing steps).

We next discuss the text2net's computational complexity. 
To analyze it, we assume the following: the text corpus $T$ is comprised of $s$ sentences. 
In terms of space, the complexity can be divided into two main parts. First, the memory needed to store the sentence being currently processed and the memory for storing the network. As the sentences can be processed in small batches, we focus on the spatial complexity of the token network. 
Let the corpus consist of $t$ tokens. In the worst case, all tokens are interconnected and the spatial complexity is quadratic $\mathcal{O}(t^{2})$. 
Due to Zipf's law 
 networks are notably smaller as each word is (mostly) connected only with a small subset of the whole vocabulary (heavy tailed node degree distribution). The approach is thus both spatially, as well as computationally efficient, and can easily scale to corpora comprised of hundreds of thousands of sentences.

\begin{algorithm}[!b]
\KwData{Text corpus $T$ (of documents $d_1 \dots d_n$), empty weighted network $G$}
\Parameter{Minimum number of tokens per sentence $t_s$, Minimum token length $t_l$, word transformation function $f$, stopwords $\sigma$, weight prunning threshold $\theta$, frequency weight function $\rho$}
\KwResult{A weighted network $G$}
\For{$d \in T$}{
    orderedTokens := getTokens($d$, $t_l$,$t_s$,$f$,$\sigma$)\Comment*[r]{Get token sequence.}
    \For{$q_i \in \textrm{orderedTokens}$}{
        arc := ($q_i$,$q_{i+1}$)\Comment*[r]{Construct an arc.}
        addToNetwork($G$, arc)\Comment*[r]{Construct the network.}
        \If{arc $\in$ current set of arcs of $G$}{
            update arc's weight via $\rho$\Comment*[r]{Update weights.}
        }
    }
}
G := prunenetwork($G$,$\theta$)\Comment*[r]{Prune the network.}
\Return{G}
\caption{text2net algorithm.}
 \label{algo:plab}
\end{algorithm}

In terms of hyperparameters, the following options are available (offering enough flexibility to model different aspects of a language, rendering text2net suitable as the initial step of multiple down-stream learning tasks):
\begin{itemize}
\item minimum sentence length considered for network construction ($t_s$),
\item minimum token length ($t_l$),
\item optional word transformation (e.g., lemmatisation) ($f$),
\item optional stopwords or punctuation to be removed ($\sigma$),
\item arc weight assignment function ($\rho$) (e.g., co-occurrence frequency),
\item a threshold for arc prunning based on weights ($\theta$).
\end{itemize}

\section{Considered network topology metrics}
\label{sec:metrics}
In this section we discuss the selected metrics that we applied to directed weighted networks. The metrics vary in their degree of computational complexity.
\begin{description}
\label{sec:infomap}
\item[Number of nodes.] The number of nodes present in a given network.
\item[Number of edges.] The number of edges in a given network.
\item [InfoMap communities.]
The InfoMap algorithm \cite{rosvall2009map} is based on the idea of minimal description length of the walks performed by a random walker traversing the network. It obtains a network partition by minimizing the description lengths of random walks, thus uncovering dense regions of a network, which represent communities. Once converged, InfoMap yields the set of a given network's nodes $N$ partitioned into a set of partitions which potentially represent functional modules of a given network.

\item [Average node degree.] How many in- and out connections a node has on average. For this metric, networks were considered as undirected. See below: 
$$ \textrm{AvgDeg} = \frac{1}{|N|}\sum_{n \in N}\textrm{deg}_{in}(n)+\textrm{deg}_{out}(n)$$.
\item [Network density.] The network density represents the percentage of theoretically possible edges. This metric is defined as:
\begin{equation*}
\textrm{Density} = \frac{|A|}{|N|(|N|-1)};
\end{equation*}
where $|A|$ is the number of arcs and $|N|$ is the number of nodes. This measure represents more coarse-grained clustering of a network.
\item [Clustering coefficient.] This coefficient is defined as the geometric average of the subnetwork edge weights:
\begin{equation*}
\textrm{ClusCoef} = \frac{1}{|N|}\sum_{u \in N}\bigg ( \frac{1}{deg(u)(deg(u)-1))}
      \sum_{vw} \sqrt[3]{(\hat{w}_{uv} \hat{w}_{uw} \hat{w}_{vw}}\bigg );
\end{equation*}
here, $\hat{w}_{vw}$ for example represents the weight of the arc between nodes $v$ and $w$. The $deg(u)$ corresponds to the $u$-th node's degree. Intuitively, this coefficient represents the number of closed node triplets w.r.t. number of all possible triplets. The higher the number, the more densely connected (clustered) the network. 
See \cite{bollobas2013modern} for detailed description of the metrics above.
\end{description}

\section{Language comparison experiments}
\label{sec:comparison}
In this section we discuss the empirical evaluation setting, where we investigated how the proposed network-based text representation and network-topology metrics can be used for the task of language comparison. We use the parallel corpus (i.e., corpus of aligned sentences across different languages) from the DGT corpus, i.e. Directorate-General for Translation  translation memory, provided by Joint Research Centre and available in OPUS \cite{tiedemann2012parallel}. 
We selected nine different languages: EN -- English,  ES -- Spanish, ET -- Estonian, FI -- Finish, LV -- Latvian, NL -- Dutch,  PR -- Portugese, SI -- Slovene, SK -- Slovak, covering languages from different historical origins and language families: Romance languages (PT, ES), Balto-Slavic languages including Slavic (SI, SK) and Baltic (LV) language examples, Germanic langauges (EN, NL), as well as Finnic languages from Uralic family (FI, ET).
The selected languages have also different typological characteristics. For example in terms of morphological typology, EN can be considered as mostly analytic, while majority of others are synthetic languages, where for example FI is considered as agglutinative, while Slavic languages are fusional as they are highly inflected.

The goal of the paper was to use the network topology metrics for langauge comparison. We considered all the pairs between the selected languages, resulting in 36 comparisons for each network-based metric. From the parallel corpus we sampled 100{,}000 sentences for each language, resulting in 900,000 sentences, which match across languages. 

From each language, we constructed a network using text2net with following parameters: the minimum number of tokens per sentence ($t_s$) was set to 3, the minimum length of a given token ($t_l$) to 1, the word transformation function transformed words to lower-case, no lemmatisation was used, and punctuation was removed. 
We defined $\rho(\textrm{arc}) = 1$.

We compared the pairs of languages
as follows. 
For each of the two languages, we transformed the text into a network. The discussed network topology metrics were computed for each of the two networks. Differences between the metrics' values are reported in tabular form (Table~\ref{tbl:first}), as well as visualized as heatmaps (Figure~\ref{fig:lang}). In the latter,
the cells are colored according to the \emph{absolute} difference in a given metric for readability purposes. 
Thus, the final result of the considered analysis are differences in a selected network topology metric. The selected  results were further visualized in Figure~\ref{fig:net}.

We used NLTK \cite{loper2002nltk} for preprocessing, Py3plex \cite{vskrlj2018py3plex}, NetworkX \cite{hagberg2008exploring}, Cytoscape \cite{smoot2010cytoscape} for network  analysis and visualizatino and Pandas  for numeric comparisions \cite{mckinney2011pandas}. Full code is available at: \url{https://github.com/SkBlaz/language-comparisons}.

While we do not have full linguistic hypotheses about the expected mapping of the linguistic characteristics and the topological metrics, we believe that the network-based comparisons should show differences between the languages. For example, the number of nodes might capture linguistic properties, such as inflectional morphology, where we could expect that morphologically rich languages would have more nodes. Number of edges might capture linguistic properties, such as the flexibility of the word order. The other measures are less intuitive 
and will be further investigated in future work. However, we believe that more complex the language (including aspects of morphology richness and word order flexibility), the richer the corresponding network's structure, while the number of connected components might offer insights into general dispersity of a given language, and could pinpoint grammatical differences if studied in more detail. Also clustering coefficient might be dependent on how fixed is the word order of a given language. 
None of the above has been systematically investigated, and the hypothesis is, that differences between languages will have high variability and show already known, as well as novel groupings of the languages.

\section{Results}
\label{sec:five}
In this section we present the results of cross-lingual comparison. The inter-language differences in tabular format are given in Table~\ref{tbl:first}. 
The measures given in the table are the differences in:
\#Nodes  --- the number of nodes, \#Edges  --- the number of edges, Mean degree  --- mean node degree, Density --- network density as defined in Section~\ref{sec:metrics}, MaxCom --- maximum community size, MeanCom --- mean community size, both computed using InfoMap communities, Clustering  --- clustering coefficient and CC ---  the number of connected components. The differences in the table are presented in L2-L1 absolute differences, while for nodes and edges we also present the differences as relative percentages of the e.g., number of nodes of the second language w.r.t the number of nodes of the first language\footnote{For nodes $N_{\textrm{diff}} = \frac{100 \cdot |N_2|}{|N_1|}$, and for edges $E_{\textrm{diff}} = \frac{100 \cdot |E_2|}{|E_1|}$; the first language's values are compared against the second language's values.}.
It can be observed that some language pairs differ substantially even if only node counts are considered, where EN-FI is the pair with the largest difference, which is not surprising. English is for example an analytical language, while Finnish agglutinative with very rich morphology. Further, some of the metrics indicate groupings, which can be further investigated using heatmaps and direct visualization of language-language links.

\begin{figure}[ht]
\centering
\resizebox{1\textwidth}{!}{
\begin{tabular}{ccc}
\subcaptionbox{Maximum community size}{\includegraphics[width = 2.5in]{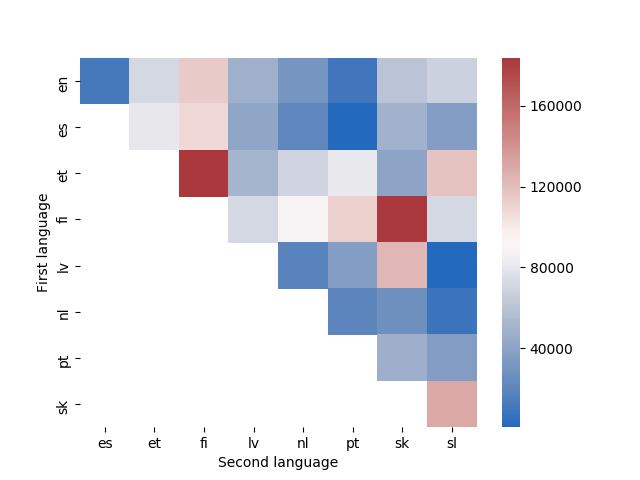}} &
\subcaptionbox{Mean community size}{\includegraphics[width = 2.5in]{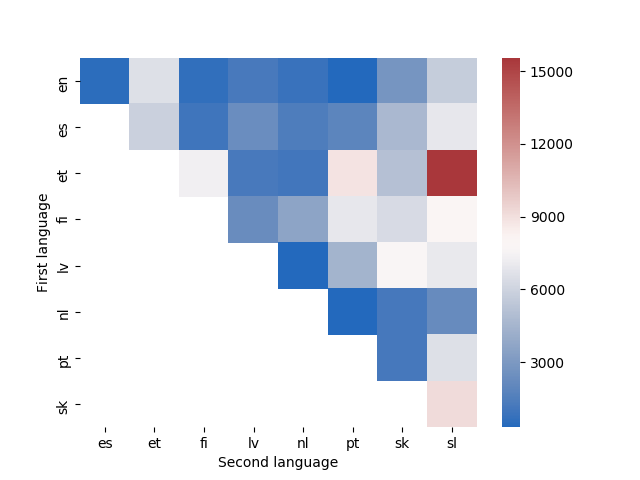}}  &
\subcaptionbox{Density}{\includegraphics[width = 2.5in]{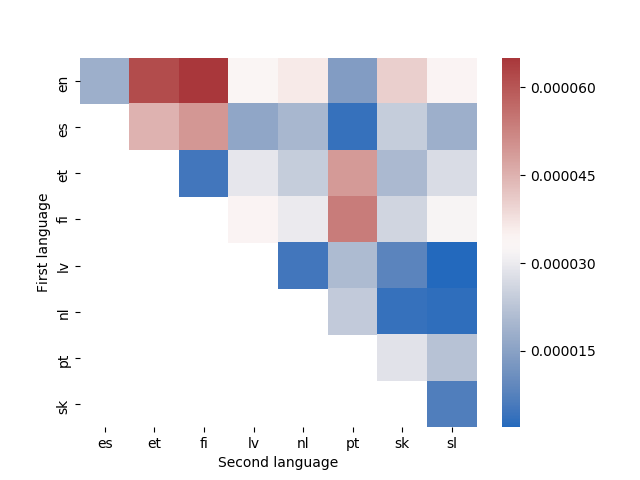}}  \\
\subcaptionbox{Average degree}{\includegraphics[width = 2.5in]{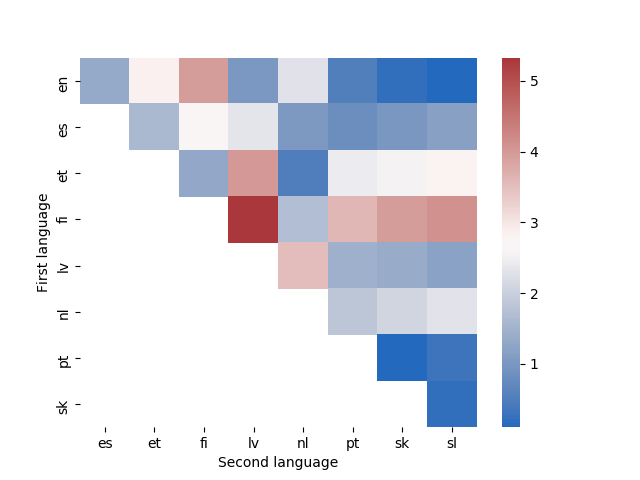}}  &
\subcaptionbox{Connected components}{\includegraphics[width = 2.5in]{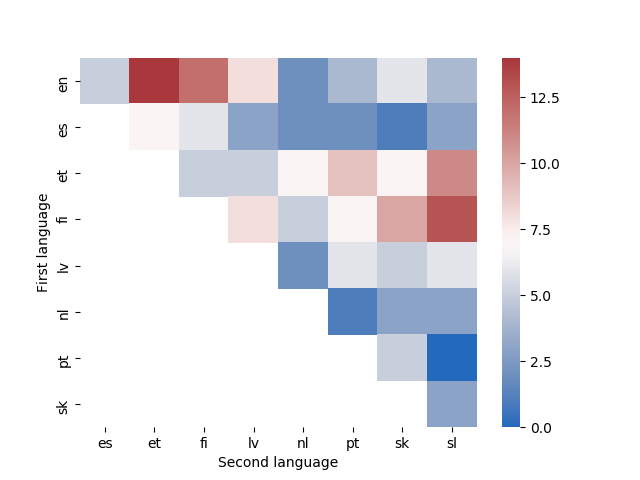}}  &
\subcaptionbox{Clustering coefficient}{\includegraphics[width = 2.5in]{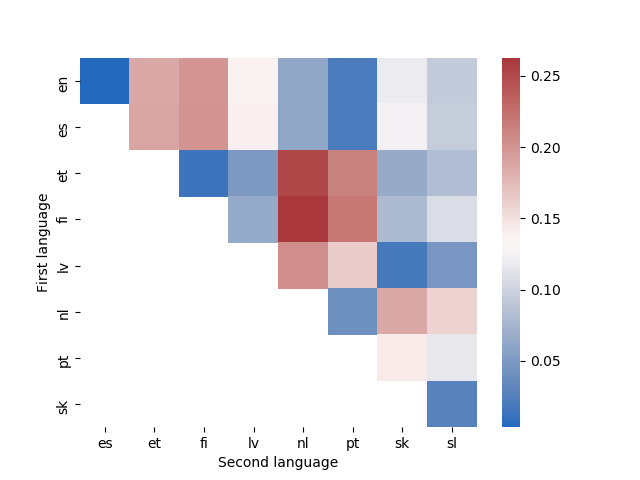}}\\
\subcaptionbox{Num. Nodes}{\includegraphics[width = 2.5in]{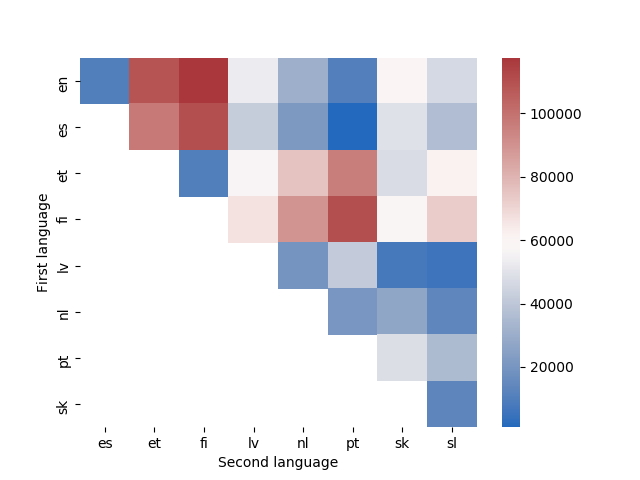}}  &
\subcaptionbox{Num. Edges}{\includegraphics[width = 2.5in]{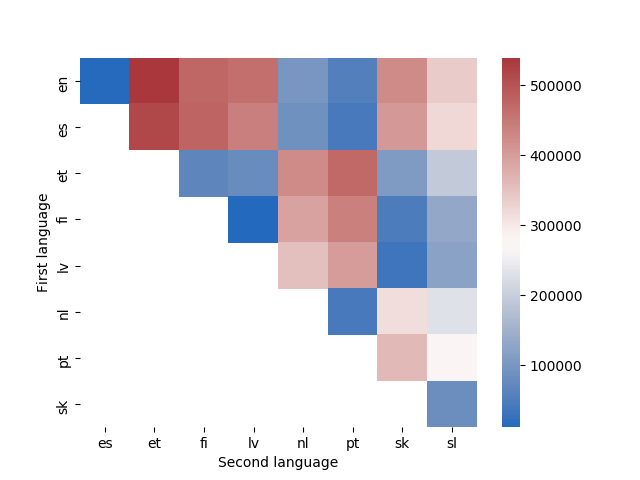}} &
\end{tabular}
}
\caption{Pairwise language comparison via various network-topological metrics. Cells represent the absolute
differences between metrics of individual text-derived networks. 
Red regions represent very different networks,
and blue very similar ones.}
\label{fig:lang}
\end{figure}

 From heatmaps shown in Figure~\ref{fig:lang}, where colors of individual cells represent differences between a given metric's values across languages, we can make several interesting observations. Based on Num. of nodes, FI and ET are very similar, and the most different to other languages. Both are agglutinative languages and part of the Uralic language family. In terms of Num. of edges, the largest differences are between ET and EN, while the most similar are LV and FI; in pairwise comparison with EN, we can see that PT, ES and NL have similar statistics, which are all languages from Germanic (NL) or Romanic family. We believe that some measures could also indicate groupings based on morphological or other typological properties beyond the currently known ones. For example, Max. community size on one hand points FI and ET as very different,  
as well as SI and SK (where in both pairs the two languages are belonging to the same language family), but on the other hand PT and ES are very similar. Further, Clustering coefficient 
 yields insights into context structure and similar properties of groupings of basic semantic units, such as words,
where high similarity between ES and PT, as well as SI and SK can be observed. Finally, the number of connected components offers insights into general dispersity of a given language, and could pinpoint grammatical differences if studied in more detail. Again, we see the most remarkable differences between EN and FI and ET, but also FI and SI, while Romanic and Germanic languages are more similar. There are many open questions. E.g., which linguistic phenomena make EN-FI being quite different in Average degree, while FI-NL are relatively similar (despite EN and NL being in the same language group)? 

\begin{figure}[t]
\centering
\includegraphics[width = .6\linewidth]{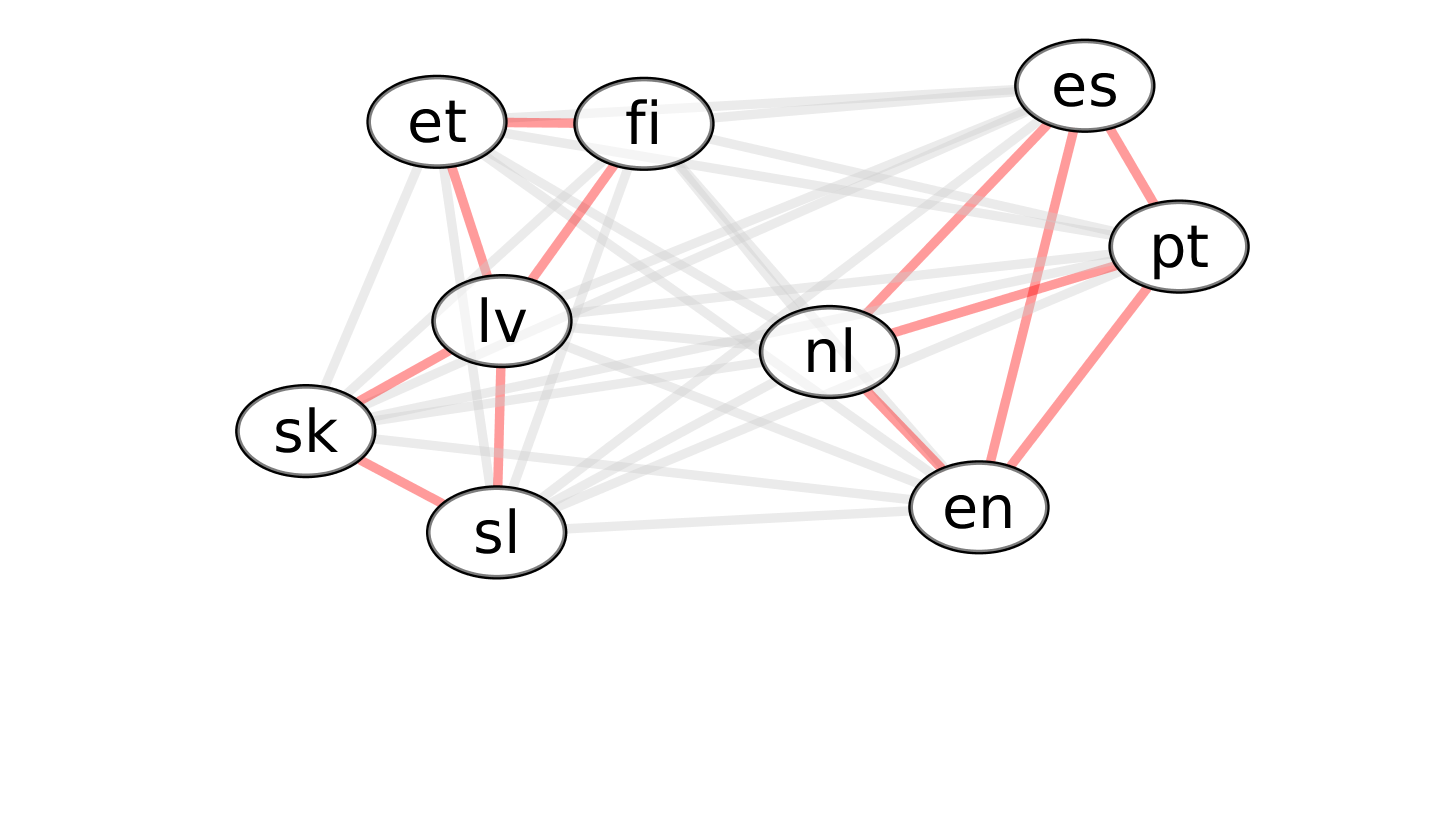}
\caption{Language network based on the Clustering coeff. The red links are present after the threshold of $10^{-3}$ was applied. Gray links represent connections that are not present given the applied threshold. We can see two groups, one formed by Balto-Slavic and Finnic languages, the other by Germanic and Romanic.}
\label{fig:net}
\end{figure}

\begin{figure}[ht]
\centering
\includegraphics[width=0.45\textwidth]{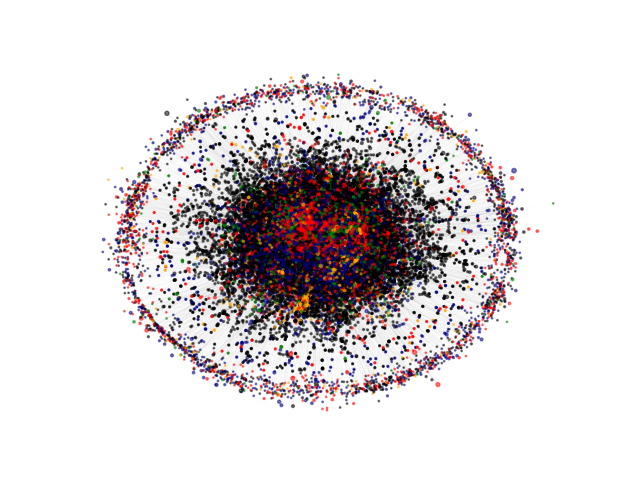}
\caption{Visualization of the English DGT subcorpus. This network was constructed using the proposed text2net algorithm, where each link corresponds to the \emph{followed by} relation between a given pair of word tokens. Clustering emerges, indicating the presence of meso-scale topological structures in such networks. Different colors correspond to different communities detected using InfoMap.}
\label{fig:corp}
\end{figure}

Clustering coefficient is also shown  in an alternative visualisation, i.e. in a colored network in Figure~\ref{fig:net}. Here, we consider Clustering coefficient metric, where we adjust the color so that it represents only very similar languages (low absolute difference in the selected metric). We selected this metric, as the heatmap
yielded the most block-alike structure, indicating strong connections between subsets of languages. We can see that Balto-Slavic and Finnic languages group together, while Germanic and Romanic form another group. Finally, we visualized the English corpus network in Figure~\ref{fig:corp}. Colored parts of the network correspond to individual communities. It can be observed that especially the central part of the network contains some well defined structures (blue and red). The figure also demonstrates, why various network-topological metrics were considered, as from the structure alone, no clear insights can be obtained at such scale.

\begin{table}[ht]
\centering
\caption{Differences between selected network-topology metrics across languages. The values are computed as L2-L1, or reported as L2 relative to L1.\vspace{0.3cm}}
\resizebox{1\textwidth}{!}{
\begin{tabular}{c||cccccccccc}

  Language pair &   \#Nodes &   \#Edges &    Mean degree &   Density ($\cdot 10^{-4}$) &  MaxCom &      MeanCom &  Clustering & CC  &  $N_{\textrm{diff}}$  (\%) &  $E_{\textrm{diff}}$ (\%) \\ \hline

 en-es &   10232 &   15251 &   -1.34 &     -0.18 &    11100 &    538.53 &                   -0.00 &                     5 &              110.90 &              101.71 \\
 en-et &  108986 &  539449 &   -2.85 &     -0.62 &   -72382 &  -6578.80 &                   -0.19 &                    14 &              233.53 &              187.45 \\
 en-fi &  117623 &  474376 &   -3.96 &     -0.65 &   114803 &    649.71 &                   -0.20 &                    12 &              249.01 &              179.60 \\
 en-lv &   53162 &  464366 &    1.02 &     -0.34 &    48411 &  -1208.95 &                   -0.14 &                     8 &              164.35 &              177.99 \\
 en-nl &   30786 &   99189 &   -2.30 &     -0.36 &    30839 &    836.35 &                    0.06 &                     2 &              138.10 &              115.73 \\
 en-pt &   10778 &   56039 &   -0.51 &     -0.14 &    10249 &   -366.93 &                    0.02 &                     4 &              113.07 &              107.48 \\
 en-sk &   59715 &  425657 &   -0.22 &     -0.41 &    60709 &   2757.51 &                   -0.12 &                     6 &              174.20 &              172.24 \\
 en-sl &   46764 &  337421 &   -0.11 &     -0.34 &   -68833 &  -5693.41 &                   -0.09 &                     4 &              156.30 &              154.82 \\
 es-et &   97822 &  518349 &   -1.62 &     -0.35&   -80506 &  -5877.34 &                   -0.19 &                     7 &              210.57 &              184.29 \\
 es-fi &  110493 &  479001 &   -2.70 &     -0.49 &   108672 &   1004.91 &                   -0.20 &                     6 &              224.53 &              176.58 \\
 es-lv &   42062 &  442253 &    2.34 &     -0.16 &    42066 &   2382.83 &                   -0.14 &                     3 &              148.20 &              174.99 \\
 es-nl &   21501 &   88846 &   -1.04 &     -0.20 &    21235 &   1433.40 &                    0.06 &                    -2 &              124.52 &              113.78 \\
 es-pt &     971 &   43922 &    0.84 &      0.04 &     1232 &   1922.25 &                    0.02 &                    -2 &              101.96 &              105.67 \\
 es-sk &   49382 &  406578 &    0.99 &     -0.20 &    49740 &   4703.95 &                   -0.12 &                     1 &              157.08 &              169.34 \\
 es-sl &   36317 &  321960 &    1.17 &     -0.18 &    36362 &   6935.34 &                   -0.10 &                    -3 &              140.94 &              152.21 \\
 et-fi &   10262 &  -68268 &   -1.32 &     -0.05 &   183810 &   7318.28 &                   -0.01 &                     5 &              106.63 &               95.82 \\
 et-lv &  -57119 &  -80883 &    4.01 &      0.29 &   -51457 &   1237.71 &                    0.05 &                    -5 &               70.38 &               94.95 \\
 et-nl &  -75247 & -424500 &    0.50 &      0.24 &   -69698 &  -1081.51 &                    0.25 &                    -7 &               59.14 &               61.74 \\
 et-pt &  -96441 & -471464 &    2.45 &      0.49 &    81260 &   8871.81 &                    0.21 &                    -9 &               48.42 &               57.34 \\
 et-sk &  -47901 & -109523 &    2.56 &      0.20 &   -40340 &   5107.84 &                    0.07 &                    -7 &               74.60 &               91.89 \\
 et-sl &  -61594 & -194218 &    2.80 &      0.27 &   117767 &  15563.93 &                    0.08 &                   -11 &               66.93 &               82.60 \\
 fi-lv &  -66730 &  -11261 &    5.33 &      0.34 &   -72108 &  -2285.02 &                    0.06 &                    -8 &               66.00 &               99.10 \\
 fi-nl &  -89718 & -393774 &    1.71 &      0.30 &   -89284 &  -3638.18 &                    0.26 &                    -5 &               55.46 &               64.44 \\
 fi-pt & -110479 & -439797 &    3.60 &      0.54 &  -111720 &  -6939.93 &                    0.22 &                    -7 &               45.41 &               59.84 \\
 fi-sk &  -59295 &  -46799 &    3.96 &      0.26 &  -182908 &  -6349.16 &                    0.08 &                   -10 &               69.96 &               95.90 \\
 fi-sl &  -72939 & -134593 &    4.11 &      0.32 &   -71835 &   8022.41 &                    0.11 &                   -13 &               62.77 &               86.20 \\
 lv-nl &  -19634 & -354516 &   -3.52 &     -0.05 &   -18654 &   -318.15 &                    0.20 &                    -2 &               84.02 &               65.02 \\
 lv-pt &  -41716 & -402441 &   -1.46 &      0.21 &   -36193 &   4468.15 &                    0.16 &                    -6 &               68.80 &               60.39 \\
 lv-sk &    7581 &  -34478 &   -1.38 &     -0.08 &  -123658 &  -7706.36 &                    0.02 &                    -5 &              105.99 &               96.77 \\
 lv-sl &   -5810 & -122602 &   -1.21 &     -0.03 &     1014 &   7032.05 &                    0.05 &                    -6 &               95.10 &               86.99 \\
 nl-pt &  -20143 &  -43781 &    1.86 &      0.24 &   -19930 &   -314.52 &                   -0.04 &                    -1 &               81.88 &               92.87 \\
 nl-sk &   27385 &  314730 &    2.09 &     -0.04 &    27329 &   1161.44 &                   -0.19 &                     3 &              126.14 &              148.83 \\
 nl-sl &   13810 &  230590 &    2.32 &      0.03&     7637 &  -2267.56 &                   -0.16 &                    -3 &              113.18 &              133.78 \\
 pt-sk &   48780 &  361817 &    0.12 &     -0.29 &    47881 &   1201.90 &                   -0.15 &                     5 &              154.06 &              160.25 \\
 pt-sl &   35260 &  275981 &    0.32 &     -0.22 &    35831 &   6622.65 &                   -0.12 &                     0 &              138.23 &              144.05 \\
 sk-sl &  -13637 &  -85421 &    0.23 &      0.07 &  -130809 &  -9182.12 &                    0.03 &                    -3 &               89.73 &               89.89 \\ \hline

\end{tabular}
}
\label{tbl:first}
\end{table}

\section{Discussion and conclusions}
\label{sec:disc}

In this work, our aim was to provide one of the first large-scale comparisons of languages based on corpus-derived networks. To the best of our knowledge, the use of network topologies on sequence-based token networks are novel and it is not yet known to what characteristics the network topologies correspond. Second, we investigated whether the difference in some metrics correspond known relationships between languages, or represent novel language groupings.

We have shown that the proposed network-based text representation offers a pallete of novel opportunities for language comparison. Commonly, methods operate on sequence level, and are as such limited to one dimensional interactions with respect to a given token. In this work we attempted to lift this constraint by introducing richer, global word neighborhood. We were able to cast the language comparison problem to comparing network topology metrics, for which we show can be informative for genetic and typographic comparisons. For example, the Slovene and Slovak languages appear to have very similar global network structure, indicating comparison using communities picks up some form of evolutionary language distance. In this work we explored only very simple language networks by performing virtually no preprocessing. We believe a similar idea could be used to form networks from lemmatized text or even Universal Dependency Tags, potentially opening another dimension.

Overall, we identified the clustering coefficient as the metric, which, when further inspected, yielded some of the well known language-language relationships, such as for example high similarity between Spanish and Portugese, as well as Slovenian and Slovak languages. Similar observation was made when community structure was compared. We believe such results demonstrate network-based language comparison represents a promising venue for scalable and more informative studies of how languages, and text in general, relate to each other.

In future, we will closer connect the interpretation of network topological features with linguistic properties, also by single language metrics. Also, we believe that document-level classification tasks can benefit from exploiting the inner document structure (e.g., the Graph Aggregator framework could be leveraged instead of/in addition to conventional RNN-based approaches). The added value of graph-based similarity for classification was demonstrated e.g., in \cite{mota2012speech} for psychosis classification from speech graphs. 
We also believe that our cross-language analysis, could be indicative for the expected quality of cross-lingual representations. 
Last but not least, we plan to perform additional experiments to see if the results are stable, leading to similar findings of other corpora genres and corpora of other sizes, and also using comparable not only parallel data.

\subsubsection{Acknowledgements}
The work was supported by the Slovenian Research Agency through a young researcher grant [B\v{S}], core research programme (P2-0103), and project Terminology and knowledge frames across languages (J6-9372). This work was supported also by the EU Horizon 2020 research and innovation programme, Grant No. 825153, EMBEDDIA (Cross-Lingual Embeddings for Less-Represented Languages in European News Media). The results of this publication reflect only the authors’ views and the EC is not responsible for any use of the information it contains.

\bibliographystyle{splncs04}
\bibliography{references}

\end{document}